\documentclass[10pt,twocolumn,letterpaper]{article}

\usepackage{wacv}              % To produce the CAMERA-READY version
\usepackage{graphicx}
\usepackage{amsmath}
\usepackage{amssymb}
\usepackage{booktabs}

\usepackage{calc}
\usepackage{subcaption}
\usepackage[american]{babel}
\usepackage{csquotes}
\usepackage{xpatch}
\usepackage{wrapfig}
\usepackage{amsmath,amsfonts}
\usepackage{algorithmic}
\usepackage{algorithm}
\usepackage{array}
\usepackage{textcomp}
\usepackage{stfloats}
\usepackage{url}
\usepackage{verbatim}
\usepackage{graphicx}
\usepackage{cite}
\usepackage{tabulary}
\usepackage{booktabs}   
\usepackage{multirow} 
\usepackage{xspace}
\usepackage{colortbl}
\definecolor{mygray}{gray}{0.9} % 这里的 0.9 表示灰色的深度。你可以调整这个值来改变颜色

\definecolor{lightRed}{RGB}{255, 25, 25}
\definecolor{darkerRed}{RGB}{125, 0, 0}
\definecolor{darkerGreen}{RGB}{20, 139, 74} 
\definecolor{lightGreen}{RGB}{0, 255, 127}

\usepackage[pagebackref,breaklinks,colorlinks]{hyperref}

\usepackage[capitalize]{cleveref}
\crefname{section}{Sec.}{Secs.}
\Crefname{section}{Section}{Sections}
\Crefname{table}{Table}{Tables}
\crefname{table}{Tab.}{Tabs.}

\def\wacvPaperID{2416} % *** Enter the WACV Paper ID here
\def\confName{WACV}
\def\confYear{2024}

\begin{document}

\title{When 3D Bounding-Box Meets SAM: \\Point Cloud Instance Segmentation with Weak-and-Noisy Supervision}
\author{Qingtao Yu$^1$, Heming Du$^1$, Chen Liu$^2$, Xin Yu$^2$ \thanks{Corresponding author}\\
$^1$College of Engineering Computting and Cybernetics,  Australian National University\\
$^2$ School of Information and Electrical Engineering, The University of Queensland\\
{\tt\small \{Terry.Yu, u6341996\}@anu.edu.au \{xin.yu, uqcliu32\}@uq.edu.au}} 

\maketitle

\begin{abstract}
Learning from bounding-boxes annotations has shown great potential in weakly-supervised 3D point cloud instance segmentation. 
However, we observed that existing methods would suffer severe performance degradation with perturbed bounding box annotations. 
To tackle this issue, we propose a complementary image prompt-induced weakly-supervised point cloud instance segmentation (CIP-WPIS) method.
CIP-WPIS leverages pretrained knowledge embedded in the 2D foundation model SAM and 3D geometric prior to achieve accurate point-wise instance labels from the bounding box annotations.  
Specifically, CP-WPIS first selects image views in which 3D candidate points of an instance are fully visible.
Then, we generate complementary background and foreground prompts from projections to obtain SAM 2D instance mask predictions. 
According to these, we assign the confidence values to points indicating the likelihood of points belonging to the instance. 
Furthermore, we utilize 3D geometric homogeneity provided by superpoints to decide the final instance label assignments. 
In this fashion, we achieve high-quality 3D point-wise instance labels.
Extensive experiments on both Scannet-v2 and S3DIS benchmarks demonstrate that our method is robust against noisy 3D bounding-box annotations and achieves state-of-the-art performance.
\end{abstract}

\section{Introduction}

Indoor point cloud instance segmentation is one of the fundamental tasks in 3D scene understanding ~\cite{schult2022mask3d, sun2022superpoint, vu2022softgroup, vu2022softgroup++, engelmann20203d, liang2021instance, liu2020learning, ngo2023isbnet, jiang2020pointgroup}. 
The goal is to predict instance masks of 3D points and corresponding semantic labels.
Current 3D indoor instance segmentation methods are mainly designed on fully-supervised annotations, \emph{i.e.}, point-wise annotation. 
However, such annotation procedures are often time-consuming and laborious due to the vast quantity of points in each scene.
Hence, there has been a growing interest in investigating weakly-supervised alternatives.
Among different types of weak supervisions for point clouds, 3D instance bounding boxes stand out as a prospective direction.
First, annotating bounding boxes is considerably efficient, as it only involves drawing a single box around each object.
More importantly, each bounding box is a naturally richer instance representation, thus making it more capable of handling instance-level segmentation.

\begin{figure*}[t]
\centering
    \begin{subfigure}[t]{0.24\textwidth}
        \centering
        \includegraphics[width=\textwidth]{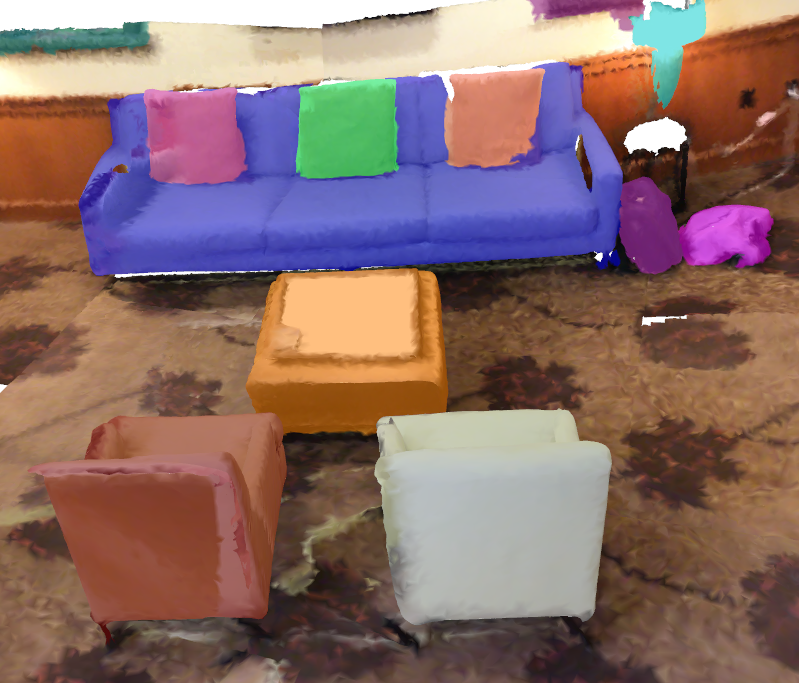}
        \caption{Point-wise annotation.}
        \label{fig:introduction-0-0}
    \end{subfigure}
    % \hspace{0.25cm}
    \begin{subfigure}[t]{0.24\textwidth}
        \centering
        \includegraphics[width=\textwidth]{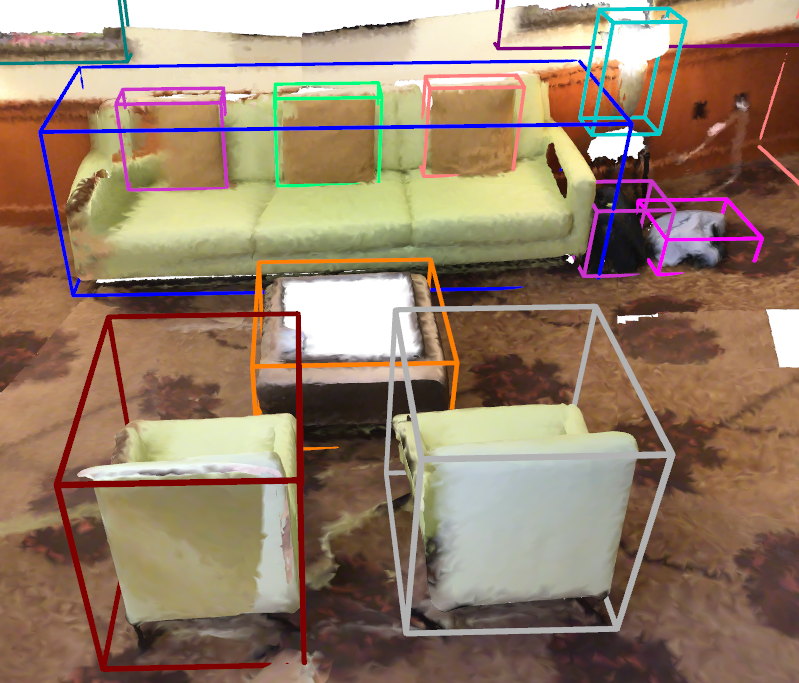}
        \caption{Bounding-box annotation.}
        \label{fig:introduction-0-1}
    \end{subfigure}
    \begin{subfigure}[t]{0.24\textwidth}
        \centering
        \includegraphics[width=\textwidth]{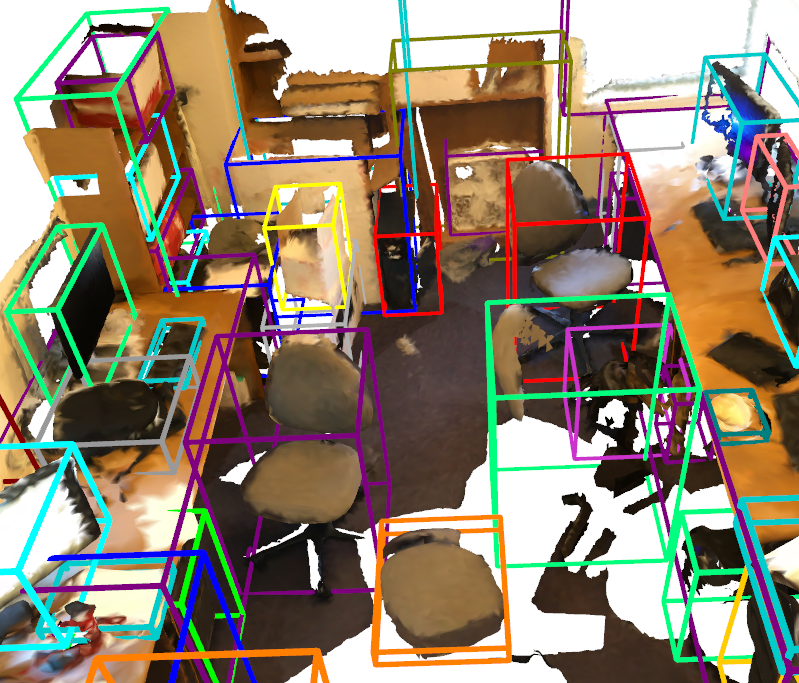}
        \caption{Indoor complex point cloud scene.}
        \label{fig:introduction-1-0}
    \end{subfigure}
    %\hspace{0.25cm}
    \begin{subfigure}[t]{0.24\textwidth}
        \centering
        \includegraphics[width=\textwidth]{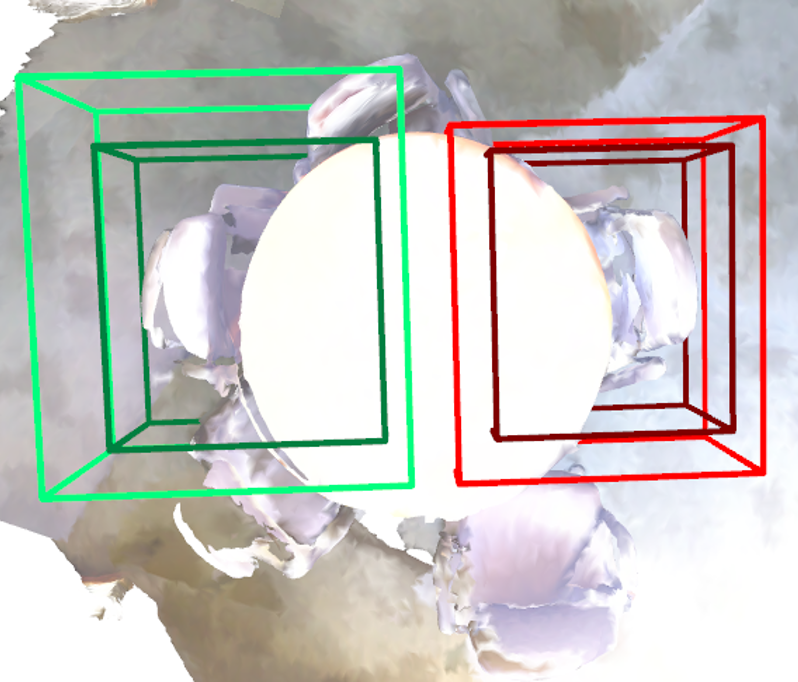}
        \caption{Noisy bounding boxes.}
        \label{fig:introduction-1-1}
    \end{subfigure}
    \caption{(a) and (b) indicate the point-wise annotation and bounding-box annotation in the point cloud instance segmentation task respectively. Compared with the point-wise annotation, labeling instances with bounding boxes notably streamline the labeling process. (c) shows the complex indoor point clouds scene, exhibiting extensive instance overlapping and highly irregular point distribution. This suggests that manual annotated bounding boxes are unavoidable suffer from inaccuracy, and thus degrade the model performance. (d) plots the examples of tight bounding-box annotations (\textcolor{darkerGreen}{$\square$} and \textcolor{darkerRed}{$\square$}) and the relaxed (\emph{i.e.} noisy) bounding boxes (\textcolor{lightGreen}{$\square$} and \textcolor{lightRed}{$\square$}). Our work aims at alleviating the negative effects brought by noisy bounding-box annotations. Experimental results (shown in Figure \ref{fig_result}) demonstrate that our method still attains better performance when utilizing the relaxed bounding boxes for supervision.}
    \label{fig:introduction}
\end{figure*}

While several 3D bounding-boxes-based instance segmentation methods have been proposed~\cite{du2023weakly,box2mask}, they all utilize the minimum axis-aligned instance bounding-boxes as annotations. In other words, these bounding boxes are the tightest ones enclosing instance point clouds along the 3D world coordination. However, in practice, manual annotations inherently contain errors or noise. 
As a result, when bounding-box annotations have minor perturbations, these methods would experience significant degradation in performance.
Therefore, it is crucial to consider the existence of noise and develop a counter-algorithm to mitigate its adverse effects. 
In this work, we proposed a complementary image prompt-induced weakly-supervised point cloud instance segmentation method under noisy bounding-box annotations. 
Our method merely requires each bounding box to cover the entire instance without any strict constraints on tightness. In other words, annotators can draw relatively looser bounding boxes as annotations, even if they are slightly larger than the objects themselves and may subsequently introduce a higher amount of noise.

As the first attempt to tackle this issue, we aim to leverage the recent advance of the 2D foundation model, \ie, Segment Anything Model (SAM)~\cite{kirillov2023segment}.
SAM performs promptable instance segmentation on the 2D domain. 
In other words, SAM cuts the objects in the image, and which object gets cut out depends on the given prompt.
Hence, we intend to generate 2D image prompts from 3D weak supervision signals and achieve accurate point-wise instance labels based on SAM predictions.
In this case, any advanced fully supervised segmentation network can be used as the following procedure.

To transfer the powerful performance of SAM on 2D domains for 3D data, we treat the points that potentially belong to the instance as candidate points and project them into multiple image planes. 
After that, a greedy view selection algorithm is employed to choose the suitable views for projection.
Based on the projected locations of candidate points, foreground 2D instance bounding boxes and sampled background pixel coordinates can be obtained as complimentary prompts.
With these prompts, we use the produced SAM predictions to effectively assign confidence values to each projected point, indicating its likelihood of belonging to the instance. 
Finally, we apply a voting scheme to uniquely assign each point to instances according to the rank of confidence.
Moreover, to mitigate the side effects of potential noisy projections and inaccurate SAM predictions, we exploit the 3D geometric structure of point clouds to facilitate our label refinement process. 
Our method does not require additional training or fine-tuning and can be adopted into any fully-supervised network. 

With our CIP-WSIS, we can accurately assign point labels even in the presence of noisy bounding-box annotations. 
Extensive experiments on both widely-used ScanNet-v2 and S3DIS benchmarks demonstrate that our method is robust against various levels of annotation noise for 3D bounding boxes. 
In particular, our CIP-WSIS outperforms the state-of-the-art Box2Mask \cite{box2mask} by a large margin
of 10\% on AP on Scannet-v2 validation set with noise-free bounding-box annotations. There is only a 2\% decrease in performance even with an increased noise rate.
Overall, our contributions are three-fold:
\begin{itemize}
    \item To the best of our knowledge, we are the first to explore the noisy weakly supervision problem on 3D instance segmentation tasks. To tackle the problem, we propose a complementary image prompt-induced weakly-supervised point cloud instance segmentation (CIP-WPIS) method that mitigates the model performance degradation problem caused by perturbed bounding box annotations.
    \item We introduce a 3D confidence ensemble module to mine the instance knowledge ensembles in the large 2D foundation model. With the 2D instance knowledge and the 3D geometric constraints, we can obtain accurate labels for each point.
    \item Our proposed is a flexible plug-in module that can be easily integrated into any fully supervised 3D instance segmentation methods, avoiding re-designing specific weak-supervised network structures. 
\end{itemize}

\section{Related Work}

\subsection{Fully-supervised 3D Instance Segmentation}

Early methods can be grouped into two classes: proposal-based and grouping-based paradigms. 
Proposal-based methods employ a top-down strategy to generate region proposals and then segment the instance within each proposal. 
For instance, \cite{yang2019learning} and ~\cite{hou20193d} first regress 3D bounding boxes for all instances and then leverage the point features to produce instance masks. 
% In~\cite{liu2020learning}, a Gaussian instance center network has been proposed to estimate instance center heatmaps. 
Grouping-based methods implement a bottom-up pipeline that produces point-wise predictions and then cluster points into different instances.
For example, MASC \cite{liu2019masc} leverages the mesh graph to cluster the instances after extracting the semantic features of points.   
To group instances more robustly, Jiang \emph{et al.} \cite{jiang2020pointgroup} proposes to estimate point offsets with respect to object centers.  
Following such design, Liang \emph{et al.} \cite{liang2021instance} and Chen \cite{Chen_2021_ICCV} improve the performance by adopting hierarchical aggregation schemes. 
Furthermore, \cite{engelmann20203d} introduces an additional graph convolutional network to refine the grouping outputs.  
Recently, SoftGroup~\cite{vu2022softgroup, vu2022softgroup++} leverages the advantages of both strategies. They proposed an architecture with bottom-up soft grouping and a subsequent top-down refinement.

State-of-the-art methods further improve 3D instance segmentation by utilizing advancements in transformers. 
The latest works, SPFormer~\cite{sun2022superpoint} and Mask3D~\cite{schult2022mask3d}, both implement a transformer decoder following the design of cutting-edge 2D segmentation methods \cite{carion2020end, cheng2021mask2former}. 
Instead of clustering points, such methods learn a set of instance queries and compute instance masks directly based on the similarities between point features and query vectors. 
Such a strategy can better model the relationship between objects and points while accelerating the inference process at the same time. 
Moreover, an attention mask mechanism is incorporated to enhance training efficiency.
However, all these fully-supervised methods require clean point-level annotations, and they would suffer severe performance drops when noisy annotations are provided.

\subsection{Weakly-supervised 3D Instance Segmentation}
Sparse-point weak supervision approaches only use a small portion of labeled point clouds to train the network.
For example, Xu \emph{et al}.~\cite{xu2020weakly} propose to propagate gradients of labeled points to those of labeled points in optimizing their semantic segmentation network.
Hou \emph{et al}.~\cite{hou2021exploring} annotate 0.1\% of points and train a 3D semantic segmentation network by contrastive learning. 
Liu \emph{et al}.~\cite{liu2021one} generate semantic pseudo labels from one point per object by implementing a contrastive learning strategy.
In addition, Wu \emph{et al}. \cite{wu2022dual} introduced a dual transformer model to effectively regularize unlabeled 3D points through an adversarial strategy at both the point level and region level.
Note that most of these works focus on semantic segmentation instead of instance segmentation. 

Compared to sparse point supervision, 3D bounding box annotations provide rough instance information such as object center and size, thus making it more capable of handling instance-level segmentation tasks. However, such a method has received little attention. 
Chibane \emph{et al}. \cite{box2mask} adopts bounding-boxes as supervision. They estimate the instance bounding boxes for each superpoint and then follow a clustering technique to decide which group of superpoints belongs to the same instance.
Du \emph{et al}.~\cite{du2023weakly} leverages 3D local geometric information to generate point-level labels from bounding-box annotations.
However, all these methods highly rely on the accuracy of bounding boxes.

\subsection{Foundation Models}
The emergence of foundation models has received lots of attention, such as \cite{devlin2018bert, radford2021learning, radford2018improving, oquab2023dinov2}. 
These models are trained on vast amounts of data and hence demonstrate superior performance. 
Very recently, the Meta Research team has released the ``Segment Anything Model" (SAM) \cite{kirillov2023segment}. It is trained on over 1 billion masks on 11 million images.
With efficient prompting, it can create high-quality, generalized masks for image instance segmentation.
Due to the excellent generalization performance of SAM, it has shown extensive use for other downstream tasks as an off-the-shelf tool \cite{huang2023instruct2act, yu2023inpaint, liu2023segment, ma2023segment, wang2023detect}. 
In our work, we aim to leverage SAM to provide 2D prior knowledge in our 3D instance segmentation.

\section{Proposed Method}

\begin{figure*}[t]
    \centering
    \includegraphics[width=\textwidth]{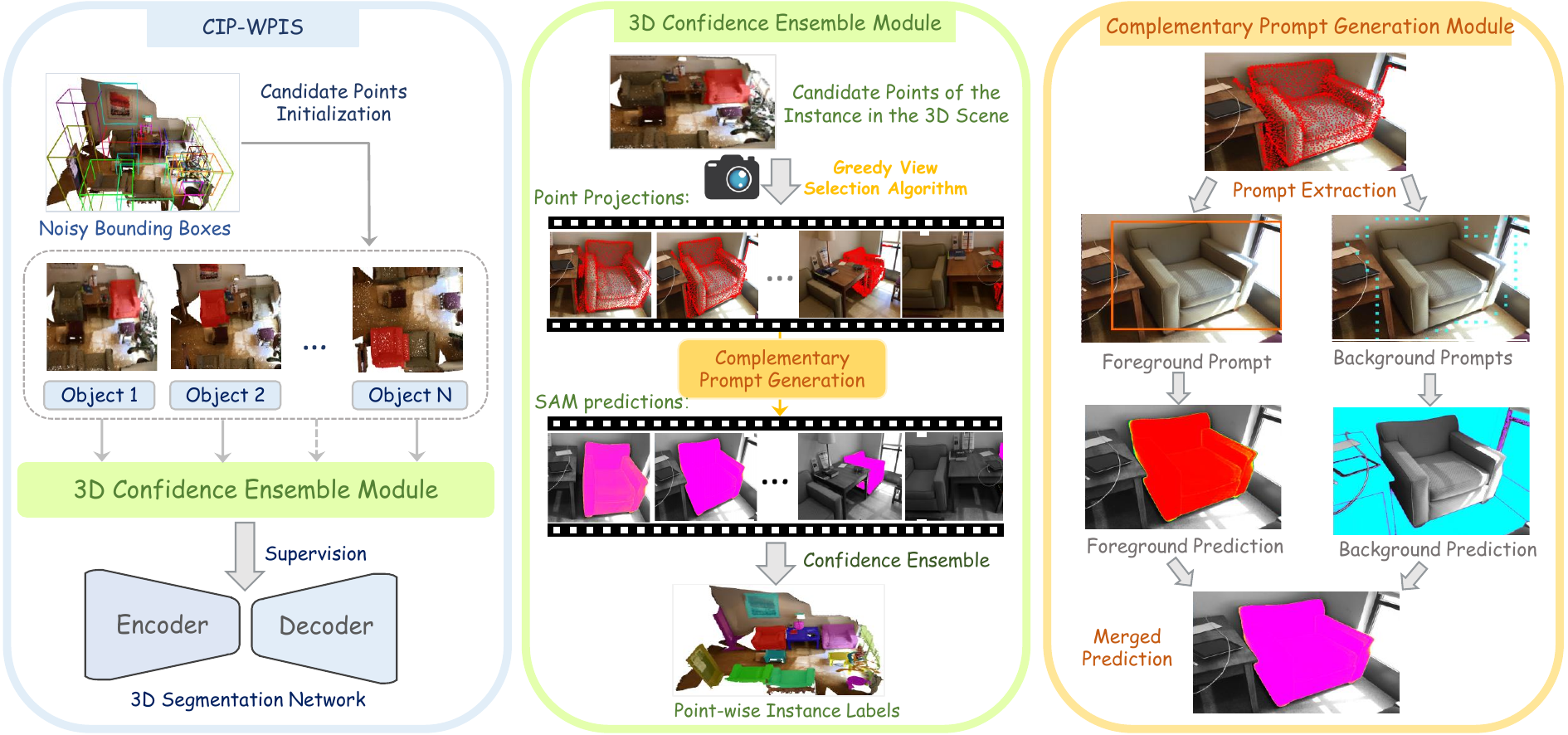}
    \caption{Workflow of our proposed CIP-WPIS. The left part depicts the whole pipeline for obtaining the point-wise instance labels from noisy bounding boxes. Specifically, we first assign the candidate points for each instance given the noisy bounding boxes. Then we devise the 3D confidence ensemble module to correct the mislabeled point of each instance. The middle part plots the concrete procedure of the 3D confidence ensemble module. To better exploit the foundation knowledge ensembled in the large 2D foundation model, we first design a greedy selection algorithm to select multiple 2D views in which an instance is fully visible. Based on projected object points in each 2D view, we introduce a complementary prompt generation module to obtain the SAM predictions from various views. After that, we integrate these predictions to indicate whether the point belongs to the instance. The right part details the complementary prompt generation module. In this module, we introduce the complementary background and foreground prompts to obtain the object mask for each instance.}

    \label{fig_pipeline}
\end{figure*}

The overall framework of our proposed CIP-WPIS is illustrated in Figure \ref{fig_pipeline}. The description of our method is detailed in the following sections. 

\subsection{Candidate Points Initialization}
\label{section1}
We aim to initialize points inside the bounding boxes that potentially belong to the corresponding instance as candidates.
Instead of considering all the contained points as candidate points, we invoke 3D superpoints to filter out some unlikely points for the efficiency of the following procedures.
Superpoints are small clusters of points symbolizing local geometric continuity, formed via a normal-based graph cluster technique \cite{felzenszwalb2004efficient}.
Following the assumption of previous works \cite{box2mask, sun2022superpoint, liang2021instance}, all the points within a superpoint belong to the same instance.
Given that bounding boxes only include additional points, we identify points as candidates only if the associated superpoints are entirely within the box.
In other words, if any point in the superpoints lies outside the box, we can confidently exclude the entire superpoint.
Following this process, we can obtain the candidate points for each instance.
Note that some points in the overlapping area of boxes might be candidate points for multiple instances, and some background points might be incorrectly identified.
The false-positive candidates will be corrected through the following processes with 2D pretrain knowledge from the fundamental model SAM.

\subsection{View Selection}
\label{section2}
Indoor 3D point clouds are reconstructed from a sequence of RGBD images.
Hence, each point of the scene is presented in at least one of the images, and this motivates us to leverage the prior knowledge from 2D for 3D tasks.
We intend to select 2D image views for instances to project all the corresponding 3D candidate points.
Due to the extensive overlapping and obstructions, it is usually difficult to locate a single instance view so that all the candidate points can be fully observed.
Therefore, we designed a greedy view selection algorithm to progressively select a subset of 2D image views for each instance.
First, we initialized an empty view set and labeled all the candidate points as unprojected. 
Then the view with the maximum number of visible points is added to the view set. 
These points, once visible, are subsequently marked as observed.
We repeat the above procedure until all the candidate points are observed.
The specifics of this algorithm are presented in Algorithm \ref{alg}.
In the projection phase, we mapped the 3D point location onto the 2D image plane utilizing the given camera view information. 
The projected 2D coordinates of each point are computed as below following the pinhole camera model:
\begin{center}
    $ \displaystyle
    \begin{bmatrix} u \\ v \\ z \end{bmatrix}
    =
    K \cdot P \cdot
    \begin{bmatrix} X \\ 1 \end{bmatrix} \quad \quad 
    \begin{bmatrix} x \\ y \end{bmatrix} 
    =
     \frac{1}{z}
    \begin{bmatrix}
    u \\ v
    \end{bmatrix}$,
  
\end{center}
where $K$ and $P$ represent the camera intrinsic and extrinsic parameter matrix. 
$X$ is the input point location vector under 3D world coordinates and $x,y$ is the projected 2D pixel coordinates.
Subsequently, we capture the points in the image view by clipping according to the dimensions of the image.
We can determine the visibility of points based on the alignment between the depth of the RGBD image and the z-axis value from the camera coordinates. %\vspace{0.5cm}

\begin{algorithm} 
\caption{Greedy View Selection}\label{alg}
\begin{algorithmic} 
\STATE \textbf{Input:} \text{Number of instances $\mathbb{N}$}  
\STATE \hspace{0.95cm} \text{Camera views set $\mathbb{V}$}
\STATE \hspace{0.95cm} \text{Candidate points sets $\mathbb{P}$ $(|\mathbb{P}|=\mathbb{N})$} 
\STATE \textbf{Output:} \text{Sets of selected instance views $V$}
\STATE \textbf{Procesure:}
\STATE \hspace{0.3cm} $ V \gets \{V_i \big\vert V_i=\emptyset , i\in \{1..\mathbb{N} \} \}$ \vspace{0.1cm} 
\STATE \hspace{0.3cm} $ S \gets \{S_j \big\vert S_j=project(\mathbb{P}, \mathbb{V}_{j}), \mathbb{V}_{j} \in \mathbb{V} \}$ \vspace{0.1cm}
\STATE \hspace{0.3cm} \textbf{for} $i=0$ \textbf{to} $\mathbb{N}$ \textbf{do}
\STATE \hspace{0.7cm} $P \gets \mathbb{P}_i$  \hspace{1.2cm} 
%\textit{\# initialise unprojected points}
\STATE \hspace{0.7cm} \textbf{while} $P \neq \emptyset$ \text{:} 
\STATE \hspace{1.1cm} $j= \arg\max \{|S_j| \big\vert S_j \in S \}$
\STATE \hspace{1.1cm} $V_i \gets V_i \cup \{\mathbb{V}_j\}$ 
\STATE \hspace{1.1cm} $P \gets P \setminus S_j$
\STATE \hspace{0.3cm} \textbf{return} $V$
\end{algorithmic} 
\label{alg1}
\end{algorithm} 

\subsection{Complementary Prompt Generation for SAM} 
\label{section3}
Once comprehensive image views for all instances within the scene are selected, we derive 2D prompts based on the projected coordinates for these views.
In this context, we adopt a complementary prompt strategy by computing a 2D bounding box of projected pixels as the foreground prompt and sampling pixels around the projected area as the background prompt.
The foreground prompt can segment the target instance within the bounding box range, and the background prompt aids in identifying irrelevant parts of the instance in the image plane.
Such dual types of prompts collectively facilitate each other to achieve the optimal 2D instance segmentation results using SAM.  

The exclusive reliance on a single type of prompt would decrease the precision of SAM predictions, primarily due to two key factors.
First, the major issue with foreground prompts is their lack of accuracy.
Since our input candidate points are highly noisy, the bounding box formed from the projected points generally exceeds the optimal size.
When there are multiple instances in the box area, SAM may over-include additional parts other than the desired instance. 
At some viewing angles, when the image only includes a small fraction of the true positives but a larger number of false positives, such a problem can be enlarged.
On the other hand, background prompts present their own set of challenges.
Compared to bounding boxes, point prompts provide weaker constraints.
Given a point as a prompt, SAM will return the most likely corresponding instance based on its pretrained knowledge.
Since SAM is a generalized model trained with millions of diverse 2D object masks, it may have a different semantic understanding from the specific dataset. 
For instance, when there is a sink within the cabinet and the sink is the target instance, SAM sometimes interprets the cabinet and sink as a singular instance if the prompt is on the cabinet.
Without a foreground prompt, the whole sink will be treated as part of the cabinet, which adversely impacts the prediction results for both classes.  

With these two types of prompts, we can leverage SAM to compute instance masks on the 2D image plane. 
Note that each mask generated by SAM is a float-type score matrix with the size of the input image.  
Then, we merge the SAM mask predictions as the following formulation and output a single mask in the view $m$ of instance $k$:
\begin{center}
    $H^m_k = H^{m,f}_k - \beta \max\{H^{m,b_1}_k,...,H^{m,b_n}_k\}$,
\end{center}
where $H^{m,f}_k$ and $H^{m,b_i}_k$ represent the mask predictions of view from the forground bounding box prompt ($f$) and background point prompts ($b$). 
In the equation, $\max$ donates the element-wise maximum operation, and $n$ is the number of background prompts. 
Since each $H^{m,b_1}_k$ only represents one instance of an irrelevant image area, we implement such a strategy to merge the background predictions as a single background mask. 
In addition, $\beta$ is a hyperparameter.
According to our ablation study, $\beta=0.5$ achieves the best performance.
In this scenario, we derive the 2D instance scores for each pixel in the selected views.
A higher score demonstrates a greater possibility that the pixel belongs to the prompt-induced instance.  

\subsection{3D Confidence Ensemble from Selected Views}
\label{section4}
Based on the obtained 2D instance prior from SAM, we designed the following function to assign the confidence values for each candidate point:
\begin{center}
   $\displaystyle C_{p,k} =\frac{ \sum _{m} \Phi( p,V_{k,m}) \cdot H^{m}_k[i,j]}{\sum _{m} \Phi( p,V_{k,m})}$.
\end{center}
The confidence result $C$ has two inputs: the 3D point $p$ and the instance ID $k$, which signifies the likelihood of the point having the corresponding instance label.
It calculates the average of the 2D mask scores of the projected pixel in the views of the instance ($V_{k,m}$). In the equation, $[i,j]$ represents the 2D projected pixel coordinate of $p$. 
Implicit function $\Phi$ stands for the visibility of point $p$ under instance view $V_{k,m}$, which outputs a value of $1$ if the point is visible in the view and $0$ if otherwise.
Note that each point might appear a different number of times across the views. Hence, a normalization factor is added below, denoting the count of visible views.  

After obtaining the confidence value of each point from 2D prior, we finalize the instance label for the candidate points guided by 3D geometric homogeneity provided by superpoints. 
We first compute the confidence of each superpoint as the mean confidence of its included points and then assign the instance labels to superpoints based on that.
A higher confidence value indicates a stronger correspondence between the superpoint and the instance.
With such a 3D cluster-level label correction approach, we can handle potential projection noise and 2D prediction errors.

We designed the following two steps to determine the instance labels of superpoints.
Firstly, we set a confidence threshold at $0$ to filter out the highly likely irrelevant superpoints from the candidate superpoints of the instance.
Despite this, there are still massive superpoints that have positive confidence values associated with multiple instances.
Therefore, we assign such superpoints to the instance with maximum confidence. 
This voting strategy aims to assign the points with the most correlated instance label. 
Following this design, additional included background points can be eliminated with negative confidence values and ambiguous points in the overlapping bounding box area can be uniquely and accurately allocated.
In this fashion, we extensively leverage 2D prior knowledge from SAM and 3D geometric information to achieve correct point-wise instance labels, which can be seamlessly integrated with any fully supervised 3D instance segmentation network.

\newcommand{\hlcell}{}

\section{Experiments}
To validate the effectiveness of our proposed CIP-WSIS, we conduct experiments on two challenging datasets, \emph{i.e.}, ScanNet-V2~\cite{dai2017scannet} and S3DIS~\cite{2017arXiv170201105A}.

\begin{table}[t]
\centering
\small
\setlength{\tabcolsep}{2.5mm}
\renewcommand{\arraystretch}{1.0}
\begin{tabular}{ m{0.8cm} m{3.2 cm} | m{0.6cm}  m{0.6cm}  m{0.6cm} }
\toprule 
Sup. & Method & $AP$ & $AP_{50}$ & $AP_{25}$ \\ \hline \hline \vspace{0.2cm}

\multirow{8}{*}{Mask} & PointGroup \cite{jiang2020pointgroup} & 0.348 & 0.517 & 0.713 \\   \vspace{0.2cm}
\quad & SSTNet \cite{liang2021instance} & 0.494 & 0.643 & 0.740 \\  \vspace{0.2cm}
\quad & SoftGroup$^{++}$ \cite{vu2022softgroup++} & 0.458 & 0.674 & 0.791 \\ \vspace{0.2cm}
\quad & ISBNet \cite{ngo2023isbnet} & 0.545 & 0.731 & 0.825 \\  \vspace{0.2cm}
\quad & Mask3D \cite{schult2022mask3d} & 0.552 & 0.737 & 0.835 \\ \vspace{0.2cm}
\quad & SPFormer \cite{sun2022superpoint} & 0.563 & 0.739 & 0.829 \\ \hline 
\vspace{0.2cm}

\multirow{2}{*}{Point} & PointContrast \cite{xie2020pointcontrast} & 0.348 & 0.517 & 0.713 \\   \vspace{0.2cm}
\quad & CSC \cite{hou2021exploring} & 0.494 & 0.643 & 0.740 \\  \hline \vspace{0.2cm}

\multirow{4}{*}{Box} & Box2Mask & 0.391 & 0.597 & 0.718 \\  \vspace{0.2cm}
\quad & WISGP \cite{du2023weakly} & 0.352 & 0.569 & 0.702 \\ \vspace{0.2cm}
\quad & \cellcolor{mygray}  Ours\&Softgroup$^{++}$ \cite{vu2022softgroup++} & \cellcolor{mygray} 0.395 & \cellcolor{mygray} 0.629 & \cellcolor{mygray}  0.773 \\\vspace{0.2cm}
\quad & \cellcolor{mygray}  Ours\&SPFormer \cite{sun2022superpoint}  & \cellcolor{mygray}  \textbf{0.475} & \cellcolor{mygray} \textbf{0.693} & \cellcolor{mygray}  \textbf{0.786} \\ \hline 
\end{tabular}
\vspace{0.3 em}
% \caption{Comparisons on \textbf{Scannet-V2 validation} set. Un}
\caption{The results of our method under noisy-free bounding boxes on the Scannet-V2 validation set. For reference purposes, we also include the results of methods using other types of supervision (Sup.), such as masks or sparse points (200 points per scene). Our method can be used as a plugin to leverage the power of a fully supervised network, and it outperforms the existing weakly supervised design. \label{tab_1}}
\end{table}

\subsection{Datasets and Evaluation Metrics}

\begin{figure*}[htbp]
    \centering
    \includegraphics[width=1.0\textwidth]{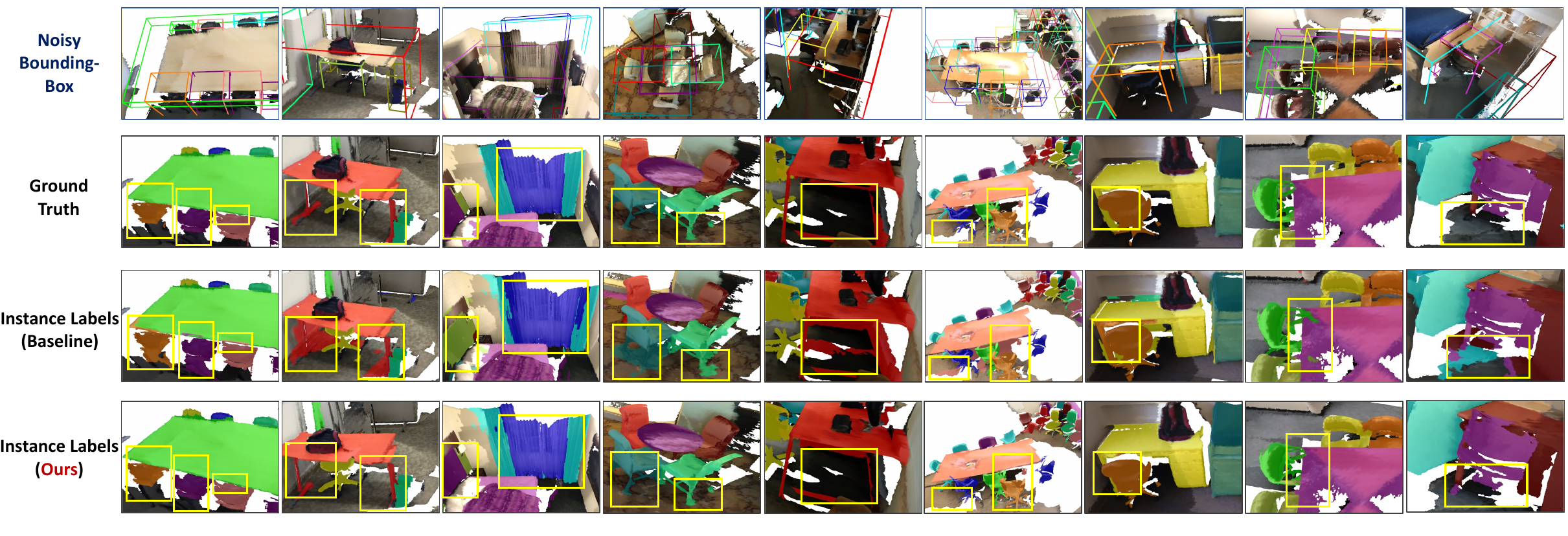}
    \caption{Visualizations of instance labels generated by the baseline model and our proposed method. To show the effectiveness of our method, we select the bounding boxes with the highest noisy rate ($\lambda=0.3$) as the input. We highlight the regions for comparison by \textcolor{yellow}{$\square$}. Even the input bounding boxes containing large unrelated regions, the instance labels generated by our method have minor differences from the ground-truth labels.}
    \label{fig_result}
\end{figure*}

\textbf{ScanNet-V2}~\cite{dai2017scannet} dataset contains 1, 613 scans with 3D semantic and instance annotations.
The dataset is split into training, validation, and testing sets, with 1, 201, 312, and 100 scans. 
It contains 18 object semantic categories. We train on the training set and report results on the validation set for comparison with other methods.
An efficient normal-based graph cut image segmentation method \cite{felzenszwalb2004efficient} is utilized for superpoint generation.
Mean Average Precision ($mAP$) serves as the common evaluation metric for instance segmentation on the Scannet-V2 dataset. It calculates the average scores of all foreground classes among different Intersection over Union (IoU) thresholds, ranging from 50\% to 95\%, with increments of 5\%. In addition, $AP_{50}$ and $AP_{25}$ represent the scores corresponding to IoU thresholds of 50\% and 25\%, respectively. We present the mAP, $AP_{50}$, and $AP_{25}$ results on the ScanNetv2 validation dataset. 

% \subsubsection{Stanford 3D Indoor Scene Dataset (S3DIS)}
\textbf{Stanford 3D Indoor Scene Dataset (S3DIS):}   S3DIS~\cite{2017arXiv170201105A} is a large-scale indoor dataset displaying six distinctive areas from three separate campus buildings. 
It contains 272 scans and is annotated with instance masks over 13 semantic classes. Following the common splits, our method is trained in Area 1, 2, 3, 4, and 6 and evaluated in Area 5.
To compare with the previous works ~\cite{du2023weakly,box2mask}, we adopt the mean precision (mPrec) and mean recall (mRec) at overlap 0.5 in S3DIS evaluation. 

\begin{table*}[t]
\centering
\small
\setlength{\tabcolsep}{2.0mm}
\renewcommand{\arraystretch}{1.0}
\begin{tabular}{m{3.7cm} | m{0.65cm} m{0.65cm} m{0.7cm} | m{0.65cm} m{0.65cm} m{0.7cm} | m{0.65cm} m{0.65cm} m{0.7cm} | m{0.65cm} m{0.65cm} m{0.65cm}  }
\hline
\multirow{3}{*}{Method} & \multicolumn{12}{c}{Noise parameter $\lambda$ }\\ 
\vspace{0.2cm}

\quad & \multicolumn{3}{c}{$\lambda=0$} & \multicolumn{3}{c}{$\lambda=0.1$} &  
\multicolumn{3}{c}{$\lambda=0.2$} & \multicolumn{3}{c}{$\lambda=0.3$} \\ \cline{2-13} 
\vspace{0.2cm}
\quad & {\small $AP$} & {\small $AP_{50}$} & {\small $AP_{25}$}  &  {\small $AP$} & {\small $AP_{50}$} & {\small $AP_{25}$} & {\small $AP$} &{\small $AP_{50}$} & {\small $AP_{25}$}  &  {\small $AP$} & {\small $AP_{50}$}& {\small $AP_{25}$} \\ 
\hline\hline
\vspace{0.2cm}
Box2Mask \cite{box2mask} & 0.398 & 0.592 & 0.712 & 0.366 & 0.563 & 0.690 & 0.359 & 0.539 & 0.670 & 0.301 & 0.515 & 0.669 \\ \vspace{0.2cm}
% WISGP \& PointGroup $^*$  & 0.313 & 0.502 & 0.649 & 0.286 & 0.472 & 0.620 & 0.261 & 0.432 & 0.583 & 0.245 & 0.417 & 0.551 \\ \vspace{0.2cm}
WISGP $^*$ \cite{du2023weakly} & 0.352 & 0.569 & 0.702& 0.333 & 0.524 & 0.645 & 0.294 & 0.492 & 0.599 & 0.261 & 0.458 & 0.562 \\ \hline \vspace{0.2cm}
% GaPro\&Softgroup \cite{ngo2023gapro} & 0.413 & 0.627 & 0.773 & - & - & - & - & - & - & - & - & - \\ \vspace{0.2cm}
% GaPro\&SPFormer \cite{ngo2023gapro}  & 0.511 & 0.704 & 0.799 & - & - & - & - & - & - & - & - & - \\ \vspace{0.2cm}
Base \& Softgroup$^{++}$ \cite{vu2022softgroup++} & 0.393 & 0.623 & 0.765 & 0.368 & 0.598 & 0.759 & 0.362 & 0.591 & 0.752 & 0.354 & 0.592 & 0.757\\ \vspace{0.2cm}
Base \& SPF \cite{sun2022superpoint} & 0.473 & 0.589 & 0.787 & 0.437 & 0.672 & 0.780 & 0.396 & 0.625 & 0.761 & 0.364 & 0.620 & 0.771 \\ \vspace{0.2cm}

\cellcolor{mygray} Ours \& Softgroup$^{++}$ & \cellcolor{mygray} 0.395 & \cellcolor{mygray} 0.629 & \cellcolor{mygray} 0.773 & \cellcolor{mygray} 0.371 & \cellcolor{mygray} 0.584 & \cellcolor{mygray} 0.741 & \cellcolor{mygray} 0.366 & \cellcolor{mygray} 0.582 & \cellcolor{mygray} 0.745 & \cellcolor{mygray} 0.364 & \cellcolor{mygray} 0.588 & \cellcolor{mygray} 0.749 \\ \vspace{0.2cm}

\cellcolor{mygray} Ours \& SPF & \cellcolor{mygray} \textbf{0.475} & \cellcolor{mygray} 0.693 & \cellcolor{mygray} 0.786 & \cellcolor{mygray} \textbf{0.465} & \cellcolor{mygray} 0.691 &  \cellcolor{mygray} 0.777 & \cellcolor{mygray} \textbf{0.452} & \cellcolor{mygray} 0.672 & \cellcolor{mygray} 0.768 & \cellcolor{mygray} \textbf{0.446} & \cellcolor{mygray} 0.668 & \cellcolor{mygray} 0.761 \\ 
\bottomrule
\end{tabular}
\caption{The results of our method under noisy bounding boxes of different noisy rates on the ScanNet-V2 validation set. The main metric for comparison is AP. Symbol $*$ indicates the method is reproduced by ourselves to test on the noisy bounding-box setting.}
\label{tab_2}
\end{table*}

\renewcommand{\arraystretch}{1.2}

\begin{table}[h]
\centering
\small
\setlength{\tabcolsep}{4.0mm}
\renewcommand{\arraystretch}{1.3}
\begin{tabular}{c| c c}
\hline
Noise & Wrong Points & Wrong Superpoints \\ 
\hline % \midrule % \vspace{0.2cm}
$\lambda=0$ & 5.2m / 3.6m  & 64k / 43k \\ % \vspace{0.2cm}
$\lambda=0.1$ & 10m / 5.9m & 114k / 67k  \\ % \vspace{0.2cm}
$\lambda=0.2$ & 13m / 7.1m   & 141k / 81k  \\ 
$\lambda=0.3$ & 16 m / 10m  & 174k / 107k  \\ 
\bottomrule
\end{tabular}
\caption{Quantitative evaluation of inaccurately assigned points and superpoints throughout the entire Scannet-V2 training set, compared between the baseline (left) and our technique (right). Due to the significant improvement in label accuracy, our method delivers a better final prediction.
\label{tab_acc}}
\end{table}

\subsection{Experiment Settings}
\textbf{Noisy Bounding-boxes:}
% \textbf{Noisy bounding-boxes settings:} 
We input noisy bounding boxes to each instance for point-wise 3D label generation. 
Each axis-aligned bounding box can be represented by a 6-dimension vector, including the $xyz$ coordinates of minimum corners and maximum corners. 
To simulate noisy annotations, we choose different hyper-parameter $\lambda$ values to enlarge the minimum bounding boxes. Each noisy bounding box can be expressed as $[C_{min}- 0.5X,  C_{max} + 0.5X]$, where $C$ stands for bounding box corners and $X = \lambda (C_{max}-C_{min})$.
To mimic human labeling in real-world settings, we add a minor Gaussian permutation with a standard deviation of $0.5\lambda X$.
A larger $\lambda$ will lead to a higher noise rate.
To exhibit robustness, we select different $\lambda$ from 0 to 0.3.

\textbf{Network Selection:} 
We selected three different seeds to generate noises and report their average performance. To evaluate the adaptive capacity of our method under different types of backbones, two representative fully supervised methods are selected (SPF and SoftGroup $^{++}$) for our experiment.
Softgroup$^{++}$ stands as a representation of the traditional methods \cite{jiang2020pointgroup, liang2021instance, engelmann20203d, vu2022softgroup}. It involves a bottom-up grouping and a top-down refinement after feature extraction.  
SPF \cite{sun2022superpoint}, similar to Mask3D\cite{schult2022mask3d}, implemented a different type of strategy using transformers \cite{vaswani2017attention} and introduces learnable queries as instance vectors. To further demonstrate the effectiveness of the method, we report the improved label accuracy in Table \ref{tab_acc}.

\subsection{Implementation Details}
\textbf{Training Strategy:} 
The fully-supervised technique usually imports a pre-train checkpoint on the same dataset with a less strong fully-supervised network backbone. To adapt to the weakly supervised scenario, we manually removed such pertaining. In this case, we make some minor adjustments to the previous training configuration, \emph{i.e.}, increasing the number of training epochs and reducing the learning rates.
To compare with the second type of baseline, we adopt the same configuration settings for training.
Moreover, \textit{ceiling} and \textit{floor} classes are considered to have the same semantic label as \textit{background} for supervision. 

\textbf{Prompt Generation:} 
Our method employs complementary prompts to guide SAM predictions. The foreground prompt is the 2D bounding box of projected pixels of candidate points. Background prompts are the sampled pixels around the projected area. Specifically, we divide the image into multiple 32 $\times$ 32 windows. If one window contains a projected point, it will be marked as `True' and vice versa. Thus, we can obtain a boolean matrix, and the number of columns and rows are image height and width divided by 32. With a simple kernel multiplication, we can achieve the windows that are close to the projected area while not including any projections. A smaller window size leads to a slightly better performance, but it will increase the number of prompts, which would cause higher computation time.

\begin{table}[t]
\centering
\small
\setlength{\tabcolsep}{3.0mm}
\renewcommand{\arraystretch}{1.2}
\begin{tabular}{ m{0.5cm} m{3.5 cm} | m{0.6cm}  m{0.6cm} }
\toprule 
{\small Sup.} & {\small Method} & {\small mPrec} & mRec  \\ \hline \hline 

\multirow{5}{*}{Mask} & PointGroup \cite{jiang2020pointgroup} & 55.3 & 42.4 \\  
\quad & SSTNet \cite{liang2021instance} & 65.5 & 64.2  \\  
\quad & SoftGroup$^{++}$ \cite{vu2022softgroup++} & 73.6 & 66.6  \\ 
\quad & Mask3D \cite{schult2022mask3d} & 68.7 & 66.3  \\ 
\quad & SPFormer \cite{sun2022superpoint} & 72.8 & 67.1 \\ \hline 
\vspace{0.1cm}

\multirow{4}{*}{Box} & Box2Mask & 66.7 & 65.5  \\  
\quad & WISGP\&PointGroup \cite{du2023weakly} & 50.0 & 52.8 \\  
\quad & WISGP\&SSTNet \cite{du2023weakly} & 44.3 & 56.7  \\ 

\quad & \cellcolor{mygray}  Ours\&SPFormer  & \cellcolor{mygray}  \textbf{69.1} & \cellcolor{mygray} 64.2 \\ \hline 
\end{tabular}
\caption{The results of our method under noisy-free bounding boxes on S3DIS folder-5. \label{tab_3}}
\end{table}

\subsection{Main Result}
The majority of the experiment is conducted on the ScanNet-V2. The quantitative results are shown in Table \ref{tab_1}, \ref{tab_2}, and qualitative results can be visualized in Figure \ref{fig_result}. 
As the first attempt to tackle the problem, we tried our best to adapt previous works and create the following two baselines for comparison. The first one is the existing box-supervised approach with the same noisy bounding boxes. Another one is state-of-the-art fully-supervised methods with point-wise labels generated from each bounding box with only the candidate point initialization step. To ensure each point has a unique instance label, we implemented a simple heuristic following the design of previous works \cite{box2mask}, which is choosing the instance with the smallest bounding box if the point is a candidate point of multiple instances. This is because smaller objects are often fully contained in bounding boxes of larger objects. 
The purpose of establishing this baseline is to ensure that the improved performance isn't just because of the change in network structure.
As a result, we achieve state-of-the-art 3D instance segmentation performance under noise-free and noisy bounding box annotations. Especially for noisy bounding boxes, our method only has around 2\% performance degradation as the noise rate increases to the next level.

We carry out extra experiments under the S3DIS dataset with SPFormer as the network structure. We achieved state-of-the-art performance in terms of mPrec under noisy-free bounding box supervision, as shown in \ref{tab_3}. For noisy bounding boxes supervision, our method suffers nearly 5\% performance drop for mPrec and 8\% for mRec with a 0.1 increase of $\lambda$. In contrast, the baseline labels with the same network structure drop 9\% and 13\%, respectively. Therefore, we prove that our CIP-WPIS is robust against noise. 

\vspace{-0.6cm}
\subsubsection{Ablation Study}
\quad \textbf{Prompt Using Strategy:} We examine another way of acquiring 2D mask scores, which is the most straightforward design. SAM allows users to feed an arbitrary number of prompts with multiple types to generate a single mask prediction. Hence, we input foreground and background prompts together to predict a single heatmap and use such heatmap to do the confidence assignment. The box foreground prompt is naturally a positive signal. Background prompts are manually set as negative signals for the predictor.  
However, we observed that when the number of prompts gets more, SAM tends to give unstable results and leads to a drop in performance.

\textbf{Hyperparameter Selection:} In addition, we evaluated different hyperparameters $\beta$ for merging the SAM predictions of each prompt. A smaller $\beta$ would be less effective in identifying the excessively included background points. And a bigger $\beta$ over-crop the true positive candidate points. The two ablation study results are combined in Table \ref{tab_4}.

\begin{table}[h]
\centering
\small
\setlength{\tabcolsep}{2.4mm}
\renewcommand{\arraystretch}{1.3}
\begin{tabular}{c | c c | c c c}
\hline
$\beta$ & Single & Merged & $mAP$ & $AP_{50}$ & $AP_{25}$ \\
\hline
0.5 & $\checkmark$  &  \:  &  0.411  & 0.628 & 0.754\\
0.2 &\: & $\checkmark$ & 0.442 & 0.681 & 0.767 \\
0.8 &\: & $\checkmark$ & 0.409 & 0.616 & 0.732 \\ 
\cellcolor{mygray} 0.5 & \cellcolor{mygray} \: & \cellcolor{mygray} $\checkmark$  & \cellcolor{mygray} \textbf{0.465} & \cellcolor{mygray} \textbf{0.691} & \cellcolor{mygray} \textbf{0.777} \\ 
\hline
\end{tabular}
\caption{ Performance comparison of different ways of leveraging prompts and different selections of hyperparameter $\beta$ under noise rate $\lambda=0.1$ with SPFormer network structure. \label{tab-a}}
\label{tab_4}
\end{table}
 
\subsection{Further Discussion}

\quad \textbf{Robustness:} We observed that our method outperforms the baseline with a bigger gap as the noise level rises. One contributing factor is the strong ability of the 2D foundation model that gives robust predictions against noisy prompts. 
Another important factor is that the number of candidate points increases along with the noise level, leading to a rise in the number of selected views. Such a consequence can be beneficial for generating a more reliable confidence value due to our averaging process. 
Therefore, in the most ideal case, if every RGBD image is taken into consideration, the labeling accuracy can be further improved. 
However, the computational burden will surge dramatically since each 3D scene may overall contain thousands of high-resolution image frames. Therefore, the greedy view selection approach is introduced as a trade-off design between final performance and computation cost. 
% \vspace{0.2cm}
 
\textbf{Limitations and Further Work:} Even though our introduced method notably enhances label accuracy, it cannot match the precision of human annotation. 
Since this work focuses on demonstrating the effectiveness of our auto-labeling module, our proposed method doesn't involve additional innovation on the fully-supervised network structure under noisy masks. 
While our method provides a confidence value for each point associated with each instance, we anticipate further studies to improve noisy bounding box supervised segmentation from a soft labeling perspective. 

\section{Conclusion}
In this work, we propose an annotation noise-aware weakly supervised point cloud instance segmentation method taking advantage of the image-domain information provided by the foundation model SAM and the geometric local consistency of point clouds.
In particular, we generate prompts on the image plane based on given weak supervision, and leverage the foundation model to mine the image-domain instance mask predictions. 
We then rectify erroneously assigned 3D point labels according to the 3D geometric consistency. 
As a result, we achieved high-quality 3D point instance labels. 
Extensive experiments demonstrate that our method outperforms the state-of-the-art, especially in the presence of noisy bounding-box annotations.

{\small
\bibliographystyle{wacv}
\bibliography{wacv}
}
\end{document}